\documentclass[runningheads]{llncs}

\usepackage{hyperref}
\usepackage{array}
\usepackage{graphicx}
\usepackage{adjustbox}
\usepackage{mathtools}
\usepackage{amsmath}
\usepackage{amssymb}
\usepackage{color}
\usepackage{xcolor}
\usepackage{caption}
\usepackage{subcaption}
\usepackage{booktabs}  
\usepackage{amsfonts}  
\usepackage{nicefrac}  
\newcommand*{\myalign}[2]{\multicolumn{1}{#1}{#2}}
\newcommand{\hide}[1]{} 
\usepackage{xspace}
\newcommand*{\eg}{\emph{e.g.},\@\xspace}
\newcommand*{\ie}{\emph{i.e.},\@\xspace}
\newcommand*{\etc}{\emph{etc.}\@\xspace}

\newcommand*{\modelname}{AnaXNet\@\xspace}

\begin{document}

\title{AnaXNet: Anatomy Aware Multi-label Finding Classification in Chest X-ray}

\author{Nkechinyere N. Agu\inst{1}\and
Joy T. Wu\inst{2} \and
Hanqing Chao\inst{1} \and \\ Ismini Lourentzou\inst{3} \and
Arjun Sharma\inst{2} \and \\
Mehdi Moradi\inst{2} \and
Pingkun Yan\inst{1} \and
James Hendler\inst{1}}
\authorrunning{N. Agu et al.}
\institute{Rensselaer Polytechnic Institute, Troy NY 12180, USA \and
IBM Research, Almaden Research Center, San Jose, CA 95120 USA \and Virginia Tech, Blacksburg, VA 24061, USA}

%
\maketitle            
\begin{abstract}
Radiologists usually observe anatomical regions of chest X-ray images as well as the overall image before making a decision. However, most existing deep learning models only look at the entire X-ray image for classification, failing to utilize important anatomical information. In this paper, we propose a novel multi-label chest X-ray classification model that accurately classifies the image finding and also localizes the findings to their correct anatomical regions. Specifically, our model consists of two modules, the detection module and the anatomical dependency module. The latter utilizes graph convolutional networks, which enable our model to learn not only the label dependency but also the relationship between the anatomical regions in the chest X-ray. We further utilize a method to efficiently create an adjacency matrix for the anatomical regions using the correlation of the label across the different regions. Detailed experiments and analysis of our results show the effectiveness of our method when compared to the current state-of-the-art multi-label chest X-ray image classification methods while also providing accurate location information.

\end{abstract}

\section{Introduction}\label{sec:intro}

Interpreting a radiology imaging exam is a complex reasoning task, where radiologists are able to integrate patient history and image features from different anatomical locations to generate the most likely diagnoses. 
Convolutional Neural Networks (CNNs) have been widely applied in earlier works in automatic Chest X-ray (CXR) interpretation, one of the most commonly requested medical imaging modality. Many of these works have framed the problem either as a multi-label abnormality classification problem \cite{rajpurkar2017chexnet,wang2017chestx}, an abnormality detection and localization problem \cite{gabruseva2020deep,sirazitdinov2019deep,wu2020automatic}, or an image-to-text report generation problem \cite{li2018hybrid,wang2018tienet}. However, these models fail to capture inter-dependencies between features or labels. Leveraging such contextual information that encodes relational information among pathologies is crucial in improving interpretability and reasoning in clinical diagnosis.

To this end, Graph Neural Networks (GNN) have surfaced as a viable solution in modeling disease co-occurrence across images. Graph Neural Networks (GNNs) learn representations of the nodes based on the graph structure and have been widely explored, from graph embedding methods \cite{grover2016node2vec,tang2015line}, generative models \cite{wang2018graphgan,you2018graphrnn} to attention-based or recurrent models \cite{li2015gated,velivckovic2017graph}, among others. For a comprehensive review on model architectures, we refer the reader to a recent survey \cite{wu2020comprehensive}. In particular, Graph Convolutional Networks (GCNs) \cite{kipf2016semi} utilize \textit{graph convolution} operations to learn representations by aggregating information from the neighborhood of a node, and have been successfully applied to CXR image classification.
For example, the multi-relational ImageGCN model learns image representations that leverage additional information from related images \cite{mao2019imagegcn}, while CheXGCN and DD-GCN incorporate label co-occurrence GCN modules to capture the correlations between labels \cite{chen2020label,liu2020dynamic}. To mitigate the issues with noise originating from background regions in related images, recent work utilizes attention mechanisms \cite{cai2018iterative,zhou2021contrast} or auxiliary tasks such as lung segmentation \cite{gordienko2018deep,chen2020two}. However, none of these works consider modeling correlations among anatomical regions and findings, \eg output the anatomical location semantics for each finding.

We propose a novel model that captures the dependencies between the anatomical regions of a chest X-ray for classification of the pathological findings, termed \textbf{Ana}tomy-aware \textbf{X}-ray \textbf{N}etwork (\modelname). We first extract the features of the anatomical regions using an object detection model. 
We develop a method to accurately capture the correlations between the various anatomical regions and learn their dependencies with a GCN model. Finally, we combine the localized region features via attention weights computed with a non-local operation \cite{wang2018non} that resembles self-attention.

The main contributions of this paper are summarized as follows: 1) we propose a novel multi-label CXR findings classification framework that integrates both global and local anatomical visual features and outputs accurate localization of clinically relevant anatomical regional levels for CXR findings, 2) we propose a method to automatically learn the correlation between the findings and the anatomical regions and 3) we conduct in-depth experimental analysis to demonstrate that our proposed \modelname model outperforms previous baselines and state-of-the-art models.
\section{Methodology}\label{sec:method}
We first describe our proposed framework for multi-label chest X-ray classification. Let CXR image collection comprised of a set of $N$ chest-X ray images $\mathcal{C} = \{{x}_1, \ldots, {x}_N \}$, where each image $x_i$ is associated with a set of $M$ labels $\mathcal{Y}_i = \{y_i^1, \ldots, y_i^M \}$, with $y_{i}^m \in \{0,1\}$ indicating whether the label for pathology $M$ appears in image $x_i$ or not. Then the goal is to design a model that predicts the label set for an unseen image as accurately as possible, by utilizing the correlation among anatomical region features $ R_i = f(x_i) \in \mathbb{R}^{k \times d}$, where $k$ is the number of anatomical region embedding, each with dimensionality $d$. The anatomical region feature extractor $f$ is described in subsection \ref{subsec:anatomical}. 
\begin{figure}[t!]
\includegraphics[width=\textwidth]{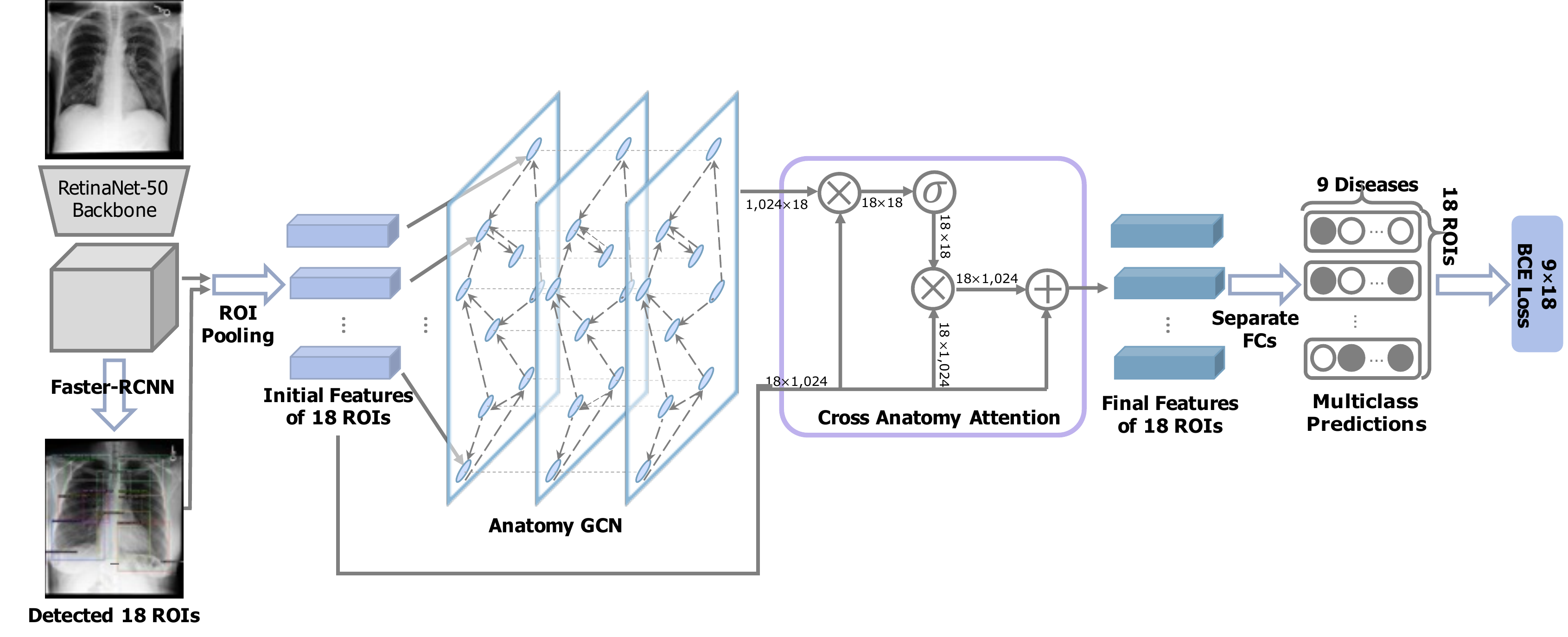}
\caption{Model overview. We extract anatomical regions of interest (ROIs) and their corresponding features, feed their vectors to a Graph Convolutional Network that learns their inter-dependencies, and combine the output with an attention mechanism, to perform the final classification with a dense layer. Note that throughout the paper, we use the term \textit{bounding box} instead of ROI.}
\label{fig:networkdiag}
\end{figure}

Given this initial set of anatomical region representations $R_i$, we define a normalized adjacency matrix $A \in \mathbb{R} ^{k \times k}$ that captures region correlations, and utilize a GCN 
$Z_i = g(R_i,A) \in \mathbb{R}^{k \times d}$ to update $R_i$ as follows:

\begin{equation}
R_i^{t+1} = \phi(AR_i^{t}W_1^{t}),
\end{equation}
where $W_1^{l} \in \mathbb{R} ^{d \times d}$ is the learned weight matrix and $\phi(.)$ denotes a non-linear operation, \eg ReLU \cite{zeiler2013rectified} in our experiments, and $t$ is the number of stacked GCN layers. 
To construct the adjacency matrix $A$, we extract co-occurrence patterns between anatomical regions for label pairs. More specifically, the label co-occurrence matrix can be computed based on Jaccard similarity:

\begin{equation}
J(r_i, r_j) = \frac{1}{M}\sum_{m=1}^{M}\frac{|\mathcal{Y}_{i}^m \cap  \mathcal{Y}_{j}^m|}{|\mathcal{Y}_{i}^m
\cup  \mathcal{Y}_{j}^m|},
\label{jaccardcoeff}
\end{equation}

where $r_{i}$ and $r_{i}$ represent anatomical regions, $\mathcal{Y}_i$ is the set for region $r_i$ and label $m$ across all images and $\cap, \cup$ denote the intersection and union over multi-sets. 
However, this label co-occurrence construction may overfit the training data due to incorporating noisy rare occurrences. To mitigate such issues, we use a filtering threshold $\tau$, \ie
\begin{equation}
A_{ij} =
    \begin{cases}
      1 & \text{if $J(R_i, R_j)$ $\geq$ $\tau$}\\
      0 & \text{if $J(R_i, R_j)$ $<$ $\tau$}
    \end{cases} ,
\end{equation}
where $A$ is the final adjacency matrix.

To capture both global and local dependencies between anatomical regions, we leverage a non-local operation that resembles self-attention \cite{wang2018non}:

\begin{equation}
Q_i = \text{softmax}\left(R_i Z_i^T\right) R_i ,
\end{equation}
where $Q_i  \in \mathbb{R}^{k \times d}$. The final prediction is computed via
\begin{equation}
\hat{y} = \left[ R_i ; Q_i \right] W_2^T,
\end{equation}
where  $W_2 \in \mathbb{R}^{2d \times M} $ is a fully connected layer to obtain the label predictions. The network is trained with a multi-label cross-entropy classification loss

\begin{equation}
L = \frac{1}{N} \sum_{i=1}^{N} \sum_{m=1}^{M} y_{i}^m log(\sigma(\hat{y}_{i}^m)) + (1-y_{i}^m)log(1-\sigma(\hat{y}_{i}^m)),
\label{losseqn}
\end{equation}
where $\sigma$ is the Sigmoid function and $\{\hat{y}_i^m, y_i^m\} \in \mathbb{R} ^{M}$ are the model prediction and ground truth for example $x_i$, respectively. The model architecture is summarized in Figure \ref{fig:networkdiag}.

\section{Experiments}\label{sec:exps}
We describe experimental details, \ie evaluation dataset, metrics, \etc, and present quantitative and qualitative results, comparing \modelname with several baselines.

\begin{table}[t!]
\caption{Dataset Characteristics. \# Images (number of images) and \# Bboxes (number of bounding boxes) labeled with L1-L9. There are a total of 217,417 images in the dataset, of which 153,333 have at least one of the L1-L9 labels globally. Of these images, 3,877,010 bounding boxes were extracted automatically and 720,098 of them have at least one or more of the 9 labels.}
 \centering
 \resizebox{0.8\textwidth}{!}{
\begin{tabular}{cccc}
 \toprule
 \textbf{{Label ID}} & \textbf{{Description}}  & \textbf{{\# Images (1)}} & \textbf{{\# Bboxes}} \\
 \midrule
L1 & Lung Opacity & 132,981 & 584,638 \\
L2 & Pleural Effusion & 68,363 & 244,005 \\
L3 & Atelectasis & 76,868 & 240,074 \\
L4 & Enlarged Cardiac Silhouette & 55,187 & 58,929 \\
L5 & Pulmonary Edema/Hazy Opacity & 33,441 & 145,965 \\
L6 & Pneumothorax & 9,341 & 22,906 \\
L7 & Consolidation & 16,855 & 53,364 \\
L8 & Fluid Overload/Heart Failure & 6,317 & 18,066 \\
L9 & Pneumonia & 32,042 & 95,215 \\
 \midrule 
All 9 labels & \textbf{{Positive/Total}} & 153,333/217,417  & 720,098/3,877,010 \\ 
\bottomrule
 \end{tabular}
}
\label{tab:dataset_description}
\end{table}

\subsection{Dataset}
Existing annotations of large-scale CXR datasets \cite{wang2017chestx,johnson2019mimic,irvin2019chexpert} are either weak global labels for 14 common CXR findings extracted from reports with Natural Language Processing (NLP) methods \cite{irvin2019chexpert}, or are manually annotated with bounding boxes for a smaller subset of images and for a limited number of labels \cite{shih2019augmenting,filice2020crowdsourcing}. None of these annotated datasets describe the anatomical location for different CXR pathologies. However, localizing pathologies to anatomy is a key aspect of radiologists' reasoning and reporting process, where knowledge of correlation between image findings and anatomy can help narrow down potential diagnoses. 
\begin{table}[t!]
\caption{Intersection over Union scores (IoU) are calculated between the automatically extracted anatomical bounding box (Bbox) regions and a set of single manual ground truth bounding boxes for 1000 CXR images. Average precision and recall across 9 CXR pathologies are shown for the NLP derived labels at: right lung (RL), right apical zone (RAZ), right upper lung zone (RULZ), right mid lung zone (RMLZ), right lower lung zone (RLLZ), right costophrenic angle (RCA), left lung (LL), left apicl zone (LAZ), left upper lung zone (LULZ), left mid lung zone (LMLZ), left lower lung zone (LLLZ), left costophrenic angle (LCA), mediastinum (Med), upper mediastinum (UMed), cardiac silhouette (CS) and trachea (Trach).}
\centering
\resizebox{\textwidth}{!}{%
\begin{tabular}{p{3cm}*{9}{p{1.3cm}<{\centering}}}
\toprule
\textbf{Bbox abbreviation} & \textbf{RL} & \textbf{RAZ} & \textbf{RULZ} & \textbf{RMLZ} & \textbf{RLLZ} & \textbf{RHS} & \textbf{RCA} & \textbf{LL} & \textbf{LAZ} \\\midrule
\textbf{Bbox IoU} & 0.994 & 0.995 & 0.995 & 0.989 & 0.984 & 0.989 & 0.974 & 0.989 & 0.995 \\
\textbf{NLP Precision} & 0.944 & 0.762 & 0.857 & 0.841 & 0.942 & 0.897 & 0.871 & 0.943 & 0.800 \\
\textbf{NLP Recall} & 0.98 & 0.889 & 0.857 & 0.746 & 0.873 & 0.955 & 0.808 & 0.982 & 1.00 \\
\midrule 
\midrule 
\textbf{Bbox abbreviation} & \textbf{LULZ} & \textbf{LMLZ} & \textbf{LLLZ} & \textbf{LHS} & \textbf{LCA} & \textbf{Med} & \textbf{UMed} & \textbf{CS} & \textbf{Trach} \\\midrule
\textbf{Bbox IoU} & 0.995 & 0.986 & 0.979 & 0.985 & 0.950 & 0.972 & 0.993 & 0.967 & 0.983 \\
\textbf{NLP Precision} & 0.714 & 0.921 & 0.936 & 0.888 & 0.899 & N/A & N/A & 0.969 & N/A \\
\textbf{NLP Recall} & 0.938 & 0.972 & 0.928 & 0.830 & 0.776 & N/A & N/A & 0.933 & N/A \\
\bottomrule
\\
\end{tabular}
}
\label{tab:dataset_validation}
\end{table}

The Chest ImaGenome dataset builds on the works of \cite{wu2020automatic,wu2020ai} to fill this gap by using a combination of rule-based text-analysis and atlas-based bounding box extraction techniques to structure the anatomies and the related pathologies from 217,417 report texts and frontal images (AP or PA view) from the MIMIC-CXR dataset \cite{johnson2019mimic}. In summary, the text pipeline \cite{wu2020ai} first sections the report and retains only the finding and impression sentences. Then it uses a prior curated CXR concept dictionary (lexicons) to identify and detect the context (negated or not) for name entities required for labeling the 18 anatomical regions and 9 CXR pathology labels from each retained sentence. The pathology labels are associated with the anatomical region described in the same sentence with a natural language parser, SpaCy \cite{spacy}, and clinical heuristics provided by a radiologist was used to correct for obvious pathology-to-anatomy assignment errors (e.g. lung opacity wrongly assigned to mediastinum). Finally the pathology label(s) for each of the 18 anatomical regions from repeated sentences are grouped to the exam level. A separate anatomy atlas-based bounding box pipeline extracts the coordinates from each frontal images for the 18 anatomical regions \cite{wu2020automatic}. 

Table \ref{tab:dataset_description} shows high-level statistics of the generated dataset. Dual annotations for 500 random reports (disagreement resolved via consensus) were curated at sentence level by a clinician and a radiologist, who also annotated the bounding boxes for 1000 frontal CXRs (single annotation). For the 9 pathology, the overall NLP average precision and recall without considering localization are 0.9819 and 0.9875, respectively. More detailed results by anatomical regions are shown in Table \ref{tab:dataset_validation}.

\subsection{Baselines}
The anatomical region feature extractor $f(x_i) \in \mathbb{R}^{k \times d}$ is a Faster R-CNN with with ResNet-50 \cite{he2016deep} as base model. Additional implementation details, \eg hyper-parameters, are provided later on (Subsection \ref{subsec:anatomical}).
We perform comprehensive analysis on the Chest ImaGenome dataset. We compare our \modelname model against:
1) \textbf{GlobalView} we implement a DenseNet169 \cite{huang2017densely} model as a baseline method to contrast the effectiveness of location-aware \modelname versus a global view of the image, 2) \textbf{Faster R-CNN} \cite{ren2015faster} followed by a fully-connected layer, \ie without the GCN and attention modules, to establish a baseline accuracy for the classification task using the extracted anatomical features, and 3) \textbf{CheXGCN} We re-implement the state-of-the-art model CheXGCN \cite{chen2020label} that utilizes GCNs to learn the label dependencies between pathologies in the X-ray images. The model uses a CNN for feature extraction and a GCN to learn the relationship between the labels via word embeddings. We replace the overall CNN with Faster R-CNN for a fair comparison with our model, but retain their label co-occurrence learning module.

\begin{table}[t!]
\centering
\caption{Comparison of our approach against baselines (AUC score).}
\resizebox{\textwidth}{!}{%
\begin{tabular}{p{3cm}*{10}{p{0.8cm}<{\centering}}}
\toprule
Method & L1 &  L2 & L3 & L4 & L5 & L6 & L7 & L8 & L9 & \textbf{AVG}  \\
\midrule
Faster R-CNN & 0.84 & 0.89 & 0.77 & 0.85 & 0.87 & 0.77 & 0.75 & 0.81 & 0.71 & 0.80\\
GlobalView & \textbf{0.91} & 0.94 & 0.86 & 0.92 & 0.92 & \textbf{0.93} & 0.86 & 0.87 & 0.84 & 0.89\\
CheXGCN & 0.86 & 0.90 & 0.91 & 0.94 & \textbf{0.95} & 0.75 & \textbf{0.89} & \textbf{0.98} & 0.88 & 0.90\\
\modelname (ours) & 0.88 & \textbf{0.96} & \textbf{0.92} & \textbf{0.99} & \textbf{0.95} & 0.80 & \textbf{0.89} & \textbf{0.98} & \textbf{0.97} & \textbf{0.93}\\
\bottomrule 
\end{tabular}
}
\label{tab:results}
\end{table}

\subsection{Implementation details}\label{subsec:anatomical}
We train the detection model to detect the 18 anatomical regions. To obtain the anatomical features, we take the final output of the model and perform non-maximum suppression for each object class using an IoU threshold. We select all the regions where any class probability exceeds our confidence threshold. We use a value of 0.5 for $\tau$. For each region, we extract a 1024 dimension convolutional feature vector. 
For multiple predictions of the same anatomical region, we select the prediction with the highest confidence score and drop the duplicates. When the model fails to detect a bounding box, we use a vector of zeros to represent the anatomical features of the region within the GCN.
We use detectron2 \footnote{\url{https://github.com/facebookresearch/detectron2}} 
to train Faster R-CNN to extract anatomical regions and their features. Our GCN model is made up of two GCN layers with output dimensionality of 512 and 1024 respectively. We train with Adam optimizer, and $10^{-4}$ learning rate for 25 epochs in total. 
\begin{table}[t!]
\begin{subtable}[t]{0.5\textwidth}
\resizebox{\textwidth}{!}{
\centering
\begin{tabular}{ccc}
\textbf{{Image 1}} & \textbf{{CS}}  & \textbf{{RCA}} \\ 
\includegraphics[width=0.4\textwidth,height=0.4\textwidth]{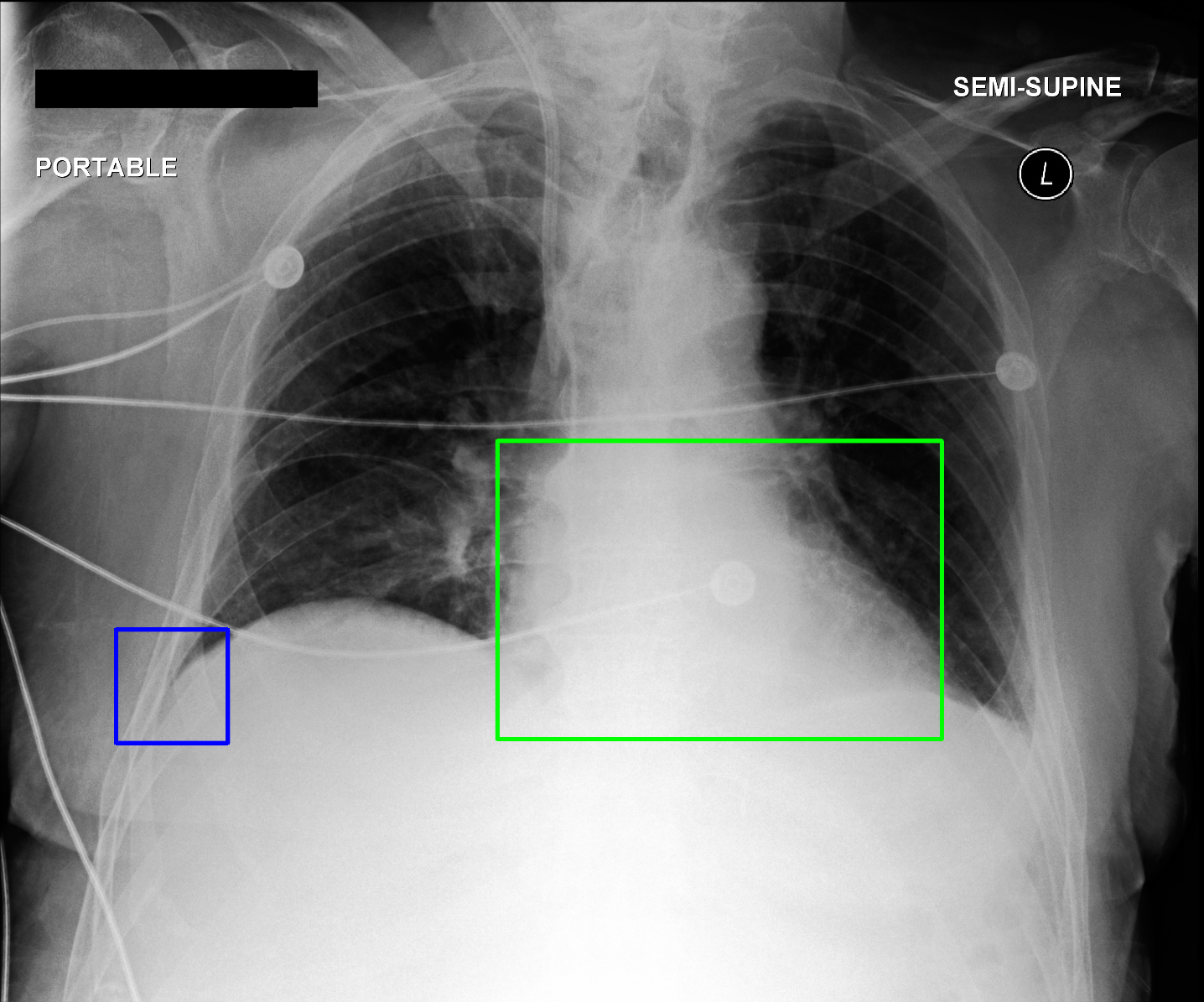} & \includegraphics[width=0.4\textwidth,height=0.4\textwidth]{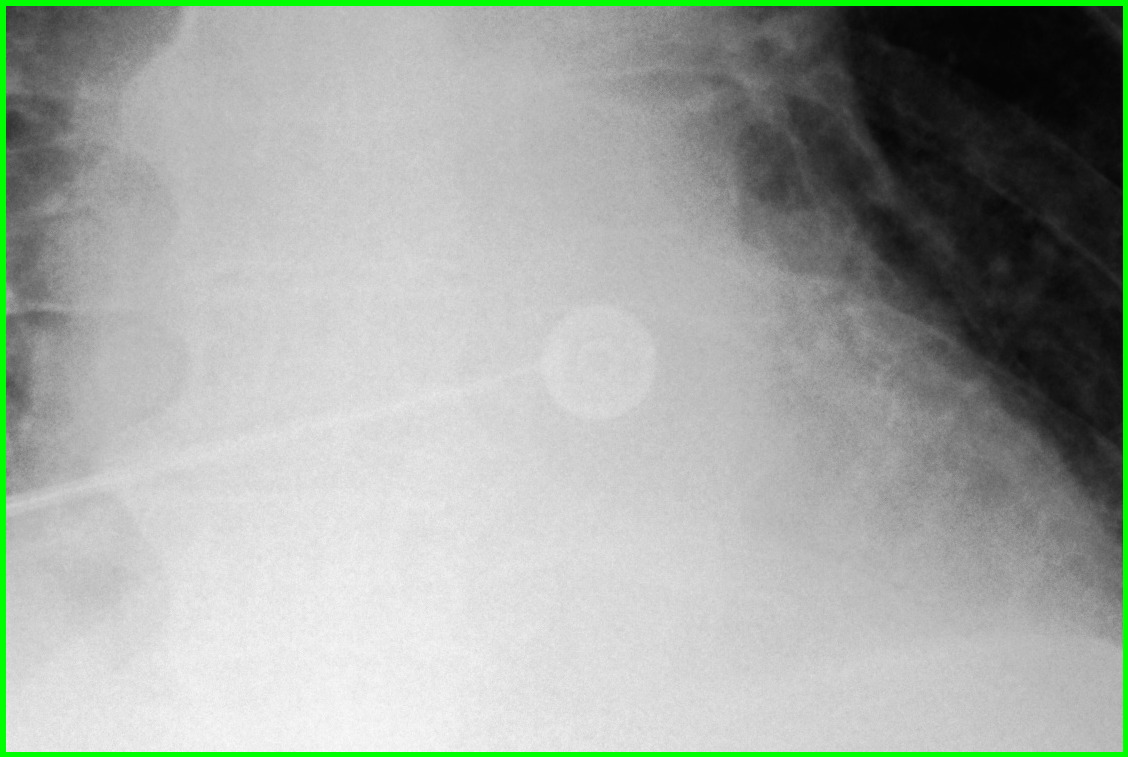} & \includegraphics[width=0.4\textwidth,height=0.4\textwidth]{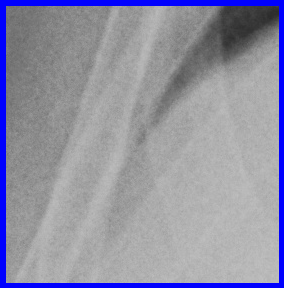} \\[0.15cm]
\myalign{l}{Ground Truth} & \myalign{l}{\textbf{No findings}} & \myalign{l}{\textbf{No findings}} \\[0.15cm] 
\myalign{l}{CheXGCN} & \myalign{l}{\color{red} \textbf{L4} } & \myalign{l}{\color{red} \textbf{L1, L2} } \\[0.15cm] 
\myalign{l}{\modelname} & \myalign{l}{\color{green} \textbf{No findings}} & \myalign{l}{\color{green} \textbf{No findings}} \\ 
\end{tabular}
}
\end{subtable} 
\hspace{0.2cm}
\begin{subtable}[t]{0.5\textwidth}
\resizebox{\textwidth}{!}{
\centering
\begin{tabular}{ccc}
\textbf{{Image 2}} & \textbf{{RCA}}  & \textbf{{LCA}} \\
\includegraphics[width=0.4\textwidth,height=0.4\textwidth]{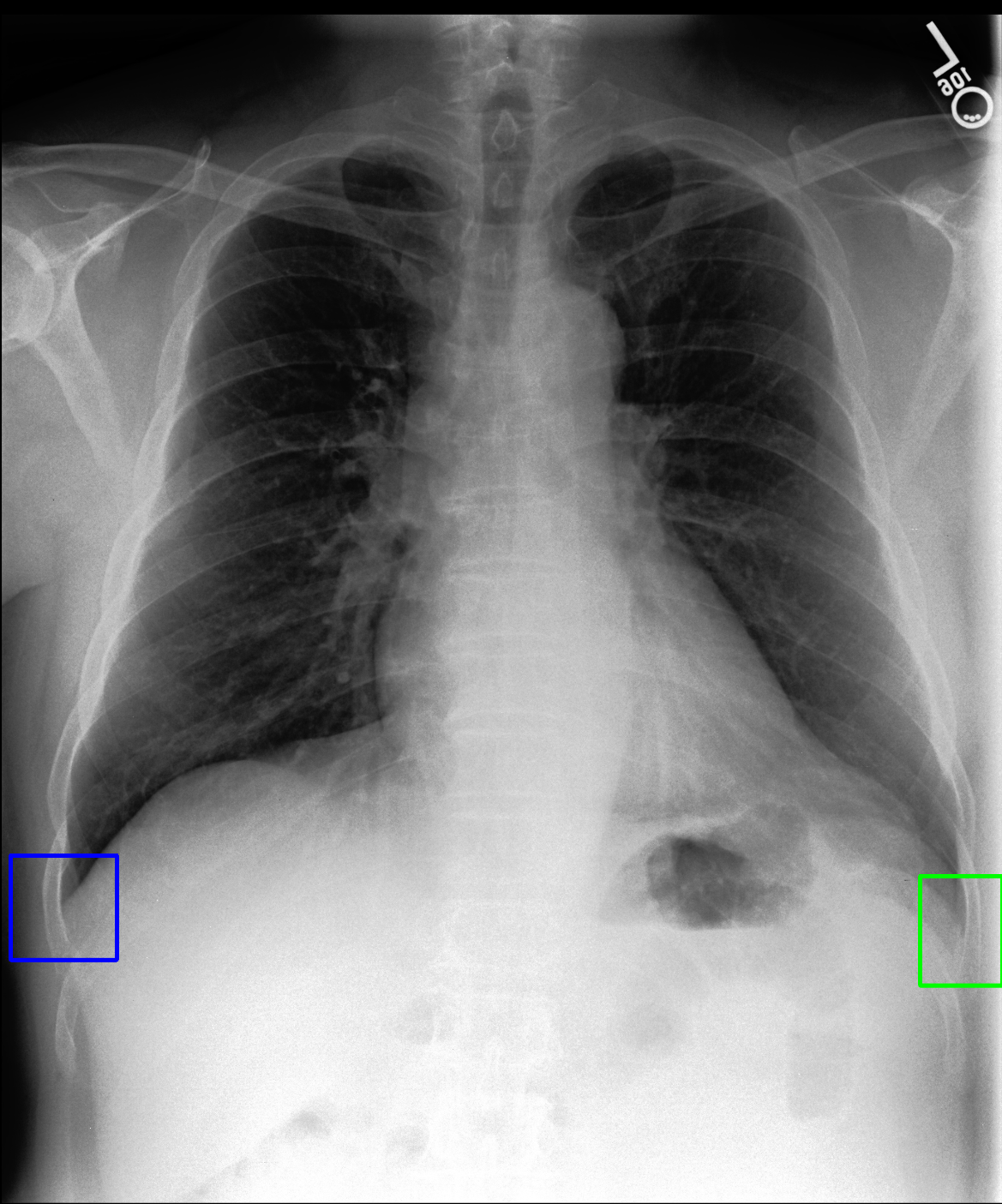} & \includegraphics[width=0.4\textwidth,height=0.4\textwidth]{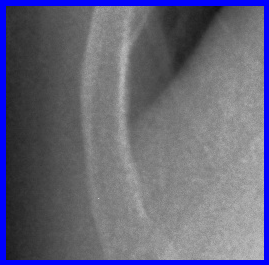} & \includegraphics[width=0.4\textwidth,height=0.4\textwidth]{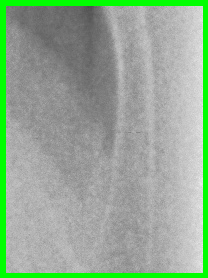} \\[0.15cm] 
\myalign{l}{Ground Truth} & \myalign{l}{\textbf{L2}} & \myalign{l}{\textbf{L2}} \\[0.15cm]  
\myalign{l}{CheXGCN} & \myalign{l}{\color{red} \textbf{No findings}} & \myalign{l}{\color{red} \textbf{No findings}} \\[0.15cm]  
\myalign{l}{\modelname} & \myalign{l}{\color{green} \textbf{L2}} & \myalign{l}{\color{green} \textbf{L2}}
\end{tabular}
}
\end{subtable}
\captionof{figure}{Examples of the results obtained by our best two models. The overall chest X-ray image is shown alongside two anatomical regions. The predictions from best performing models are compared against the ground-truth labels.
} 
\label{tab:findings}
\end{table}

\subsection{Results and Evaluation}
Results are summarized in Table \ref{tab:results}. The evaluation metric is Area Under the Curve (AUC). Note that the baseline GlobalView is in fact a global classifier and does not produce a localized label. The remaining rows in Table \ref{tab:results} show localized label accuracy. For the localized methods, the reported numbers represent the average AUC of the model for each label over the various anatomical regions. If a finding is detected at the wrong anatomical location, it counts as false detection. For fair comparison, we use the same 70/10/20 train/validation/testing split across patients to train each model. 
\modelname model obtains improvements over the previous methods while also localizing the diseases in their correct anatomical region. The GlobalView is most likely limited because it focuses on the entire image instead of a specific region. 

The CheXGCN model outperforms the other two baselines but is also limited because it focuses on one section and uses label dependencies to learn the relationship between the labels, while ignoring the relationships between the anatomical regions of the chest X-ray image.
In Table \ref{tab:findings}, we visualize the output from both the  CheXGCN model and our \modelname model. The CheXGCN model had difficulty predicting small anatomical regions like the costophrenic angles, while our model had additional information from the remaining anatomical regions, which helped in its prediction. Also the CheXGCN model struggled with enlarged cardiac silhouette label because information from the surrounding labels is needed in order to accurately tell if the heart is enlarged. 

In Figure \ref{tab:gradcam} we also visualize the output of Grad-CAM \cite{selvaraju2017grad} method on the GlobalView model to highlight the importance of the localization, while the prediction of Enlarged Cardiac Silhouette was correct, the GlobalView model was focused on the lungs. Our method was able to provide accurate localization information as well as the finding.

\section{Conclusion}\label{sec:concl}
We described a methodology for localized detection of diseases in chest X-ray images. Both the algorithmic framework of this work, and the dataset of images labeled for pathologies in the semantically labeled bounding boxes are important contributions. For our \modelname design, a Faster R-CNN architecture detects the bounding boxes and embeds them. The resulting embedded vectors are then used as input to a GCN and an attention block that learn representations by aggregating information from the neighboring regions. 

\begin{figure}[t!]
  \centering
   \resizebox{\textwidth}{!}{
  \subfloat[Original Image]{
  \includegraphics[width=0.3\textwidth, height=0.3\textwidth]{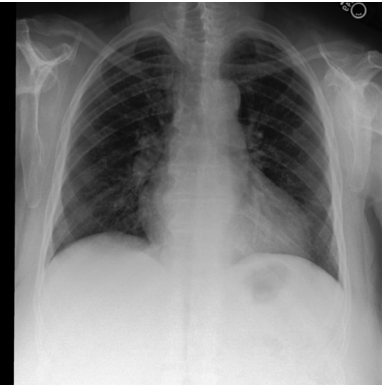}
  \label{fig:f1}}
  \hfill
  \subfloat[GlobalView \scriptsize{(Grad-CAM)}]{
  \includegraphics[width=0.3\textwidth, height=0.3\textwidth]{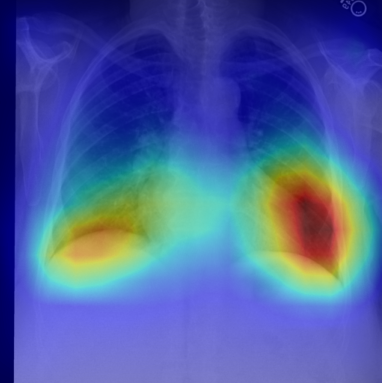}  \label{fig:f2}}
  \hfill
  \subfloat[\modelname]{
  \includegraphics[width=0.3\textwidth, height=0.3\textwidth]{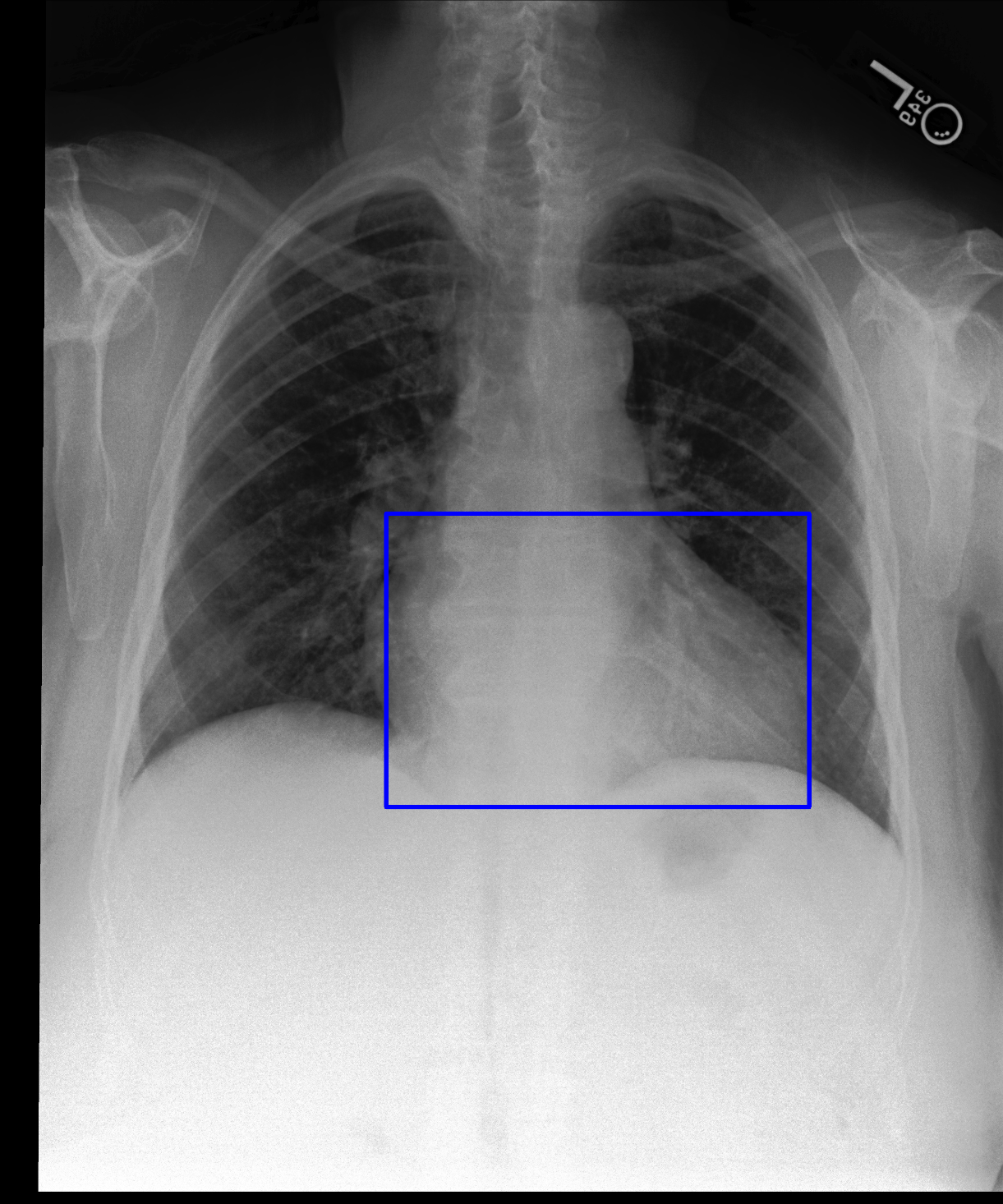} \label{fig:f3}}
  }
  \caption{An example image from our dataset with enlarged cardiac silhouette. The GlobalView network detects this label correctly, but as the Grad-CAM activation map shows (b), the attention of the network is not on the cardiac region. Our method detects the finding in the correct bounding box (c).}
  \label{tab:gradcam}
\end{figure}
This approach accurately detects any of the nine studied abnormalities and places it in the correct bounding box in the image. The 18 pre-specified bounding boxes are devised to map to the anatomical areas often described by radiologists in chest X-ray reports. As a result, our method provides all the necessary components for composing a structured report. Our vision is that the output of our trained model, subject to expansion of the number and variety of findings, will provide both the finding and the anatomical location information for both downstream report generation and other reasoning tasks. Despite the difficulty of localized disease detection, our method outperforms a global classifier. As our data shows (See Figure \ref{tab:gradcam}), global classification can be unreliable even when the label is correct as the classifier might find the correct label for the wrong reason at an irrelevant spot.

\section{Acknowledgements}
This work was supported by the 
\href{http://airc.rpi.edu}{Rensselaer-IBM AI
Research Collaboration}, part of the \href{http://ibm.biz/AIHorizons}{IBM AI Horizons Network}.

\bibliographystyle{splncs04}
\bibliography{bibliography.bib}

\end{document}